\documentclass[letterpaper, 10 pt, conference]{ieeeconf}

\IEEEoverridecommandlockouts                              

\usepackage[T1]{fontenc}

\usepackage{tikz} 

\usepackage{amsfonts}	
\usepackage{amsmath}	
\usepackage{amssymb}    
\usepackage{siunitx}
\usepackage{pifont}   

\usepackage{booktabs}
\usepackage{makecell}  
\usepackage[flushleft]{threeparttable}  
\usepackage{multirow}
\usepackage{rotating}
\usepackage{cite}

\usepackage[compact]{titlesec}  
\usepackage{xspace}    
\usepackage{xcolor}    
\let\labelindent\relax
\usepackage[inline]{enumitem} 
\usepackage{textcomp}  
\usepackage{gensymb}   
\usepackage{graphicx} 
\usepackage[caption=false,font=footnotesize]{subfig}
\usepackage{microtype}

\makeatletter
\let\NAT@parse\undefined
\makeatother

\usepackage{url}
\usepackage[pdfencoding=auto, hidelinks]{hyperref} 
\usepackage[hang,flushmargin]{footmisc}

\usepackage[capitalize]{cleveref}
\crefname{section}{Sec.}{Secs.}
\Crefname{section}{Section}{Sections}
\Crefname{table}{Table}{Tables}
\crefname{table}{Tab.}{Tabs.}

\usepackage[nogroupskip,acronyms,nopostdot,style=super,nonumberlist,toc]{glossaries}

\usepackage{flushend} 

\newacronym{ad}{AD}{automated driving}
\newacronym{hd}{HD}{high-definition}
\newacronym{fv}{FV}{frontal view}
\newacronym{bev}{BEV}{bird's-eye-view}
\newacronym{vi}{VI}{visual-inertial}
\newacronym{pgo}{PGO}{pose graph optimization}

\begin{document}

\title{\LARGE \bf
Compositional Servoing by Recombining Demonstrations
}

\author{
Max Argus*,
Abhijeet Nayak*,
Martin Büchner,
Silvio Galesso,
Abhinav Valada,
and Thomas Brox
\thanks{* Denotes equal contribution.} %
\thanks{Department of Computer Science, University of Freiburg, Germany.} 
\thanks{This work was funded by the German Research Foundation (DFG): 417962828, 401269959, the Carl Zeiss Foundation with the ReScaLe project and the BrainLinks-BrainTools center of the University of Freiburg.}
}


\maketitle


\begin{abstract}
Learning-based manipulation policies from image inputs often show weak task transfer capabilities. In contrast, visual servoing methods allow efficient task transfer in high-precision scenarios while requiring only a few demonstrations. 
In this work, we present a framework that formulates the visual servoing task as graph traversal.
Our method not only extends the robustness of visual servoing, but also enables multitask capability based on a few task-specific demonstrations. We construct demonstration graphs by splitting existing demonstrations and recombining them. In order to traverse the demonstration graph in the inference case, we utilize a similarity function that helps select the best demonstration for a specific task. This enables us to compute the shortest path through the graph. Ultimately, we show that recombining demonstrations leads to higher task-respective success. We present extensive simulation and real-world experimental results that demonstrate the efficacy of our approach.



\end{abstract}
\glsresetall




\section{Introduction}
\label{sec:intro}


The innate ability of humans to immediately copy or imitate the behavior of others has produced great interest in robotics~\cite{breazeal2002robots}. Equipping robots with these abilities leads to either requiring large computing resources and significant amounts of instructional data or to weak task transfer capability for the sake of high precision. As a consequence, enabling robots to perform complex manipulation tasks via imitation given only a few demonstrations is extremely challenging \cite{celemin2022interactive}. While it is generally possible to imitate a single specifically defined task from just one demonstration \cite{Valassakis2022}, imitating a multitude of tasks under varying circumstances represents a considerable challenge. This demands an immense effort regarding demonstration collection that is hard to automate. 
A significant share of methods approach the task of imitation learning via deep policy learning, often utilizing raw image inputs \cite{zhang2018deep, finn2017one}. Consequently, these approaches come with the drawbacks of requiring large amounts of recorded data while still exhibiting limited generalization to unseen observations. These methods are prone to failure under illumination changes, different viewpoints, or unseen scene compositions.

In this paper, we approach the problem of imitation by advocating the efficient combination of partial demonstrations to better approximate the multi-modality of diverse scenes and tasks. 
Firstly, we opt for visual servoing \cite{Mitsuda99, Argus2020, Valassakis2022}, which is able to effectively follow a demonstration to complete a task. Secondly, we observe that retrieval-based methods of different kinds perform well in a wide range of tasks where data efficiency is important, such as few-shot and long tail classification \cite{Parisot2021LongtailRV}, out-of-distribution detection \cite{sun2022knn}, 3D reconstruction \cite{what3d}, and image synthesis \cite{Qi2018SemiParametricIS,Ashual2022KNNDiffusionIG}. We draw inspiration from these methods and frame the visual servoing task as a high-level information retrieval problem. Because imitating whole demonstration trajectories does not scale well in multi-modal scenarios \cite{Izquierdo2022}, we \textit{segment} given demonstrations into parts that we call partial demonstrations. Each part handles a certain sub-task or primitive that is important to achieve the higher-level task. Specifically, we aim at recombining demonstrations that contain an initial state as well as a final state and thus address the servoing problem as a global graph optimization task.

\begin{figure}[!t]
\centering
 \centering
 \includegraphics[width=.8\linewidth, trim={0.0cm 0.0cm 0cm 0},clip]{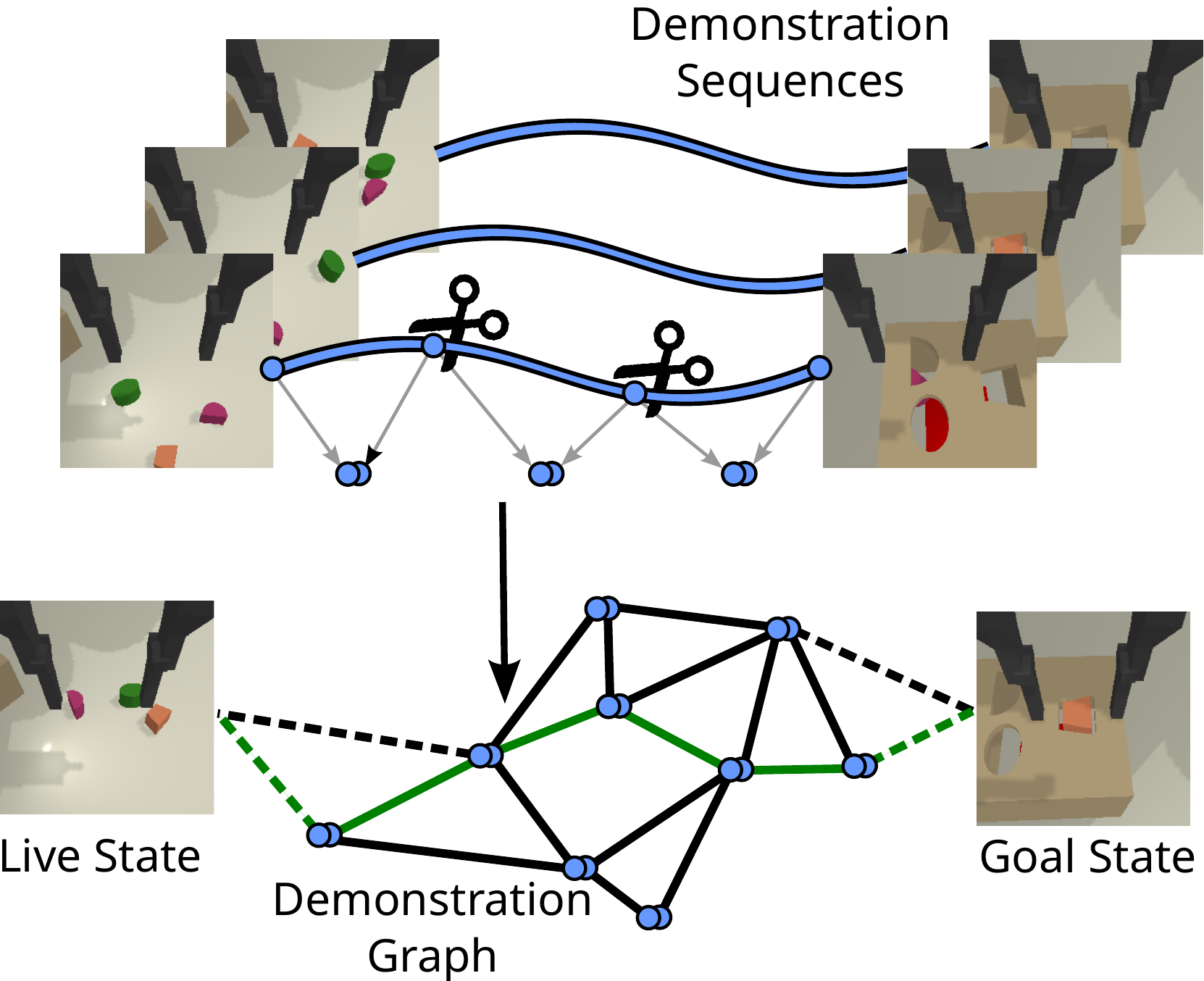}
 \vspace{-0.2cm}
 \caption{A set of demonstrations is cut into parts, which are each characterized by their beginning and end states. These parts are connected in a directed graph structure, with edges based on the similarity between connected states. During inference, we find an optimized path between a current live state and a given goal state, shown by the green lines.
 \vspace{-0.2cm}
} \label{fig:teaser}
\end{figure}

We present a retrieval-based system for few-shot imitation based on \textit{demonstration graphs} that aims to preserve the data efficiency of its original counterpart, visual servoing. In general, we argue that a wide variety of manipulation tasks can be approached by recombining parts of demonstrations, each representing different primitives under varying operating environments. In order to enable recombination via retrieval, we arrange all partial demonstrations on a demonstration graph, as shown in \cref{fig:teaser}, that is traversed in order to reach a given goal state. For example, adding a new task essentially translates to introducing a new \textit{path} on the existing graph. In general, our method implements hierarchical decision-making in the sense of high-level demonstration selection and low-level servoing based on the chosen part. The high-level planning step involves the search of demonstration graph traversals that are likely to guide the manipulator to the given goal state. The low-level operation involves the actual servoing in which a partial demonstration trajectory is imitated as closely as possible. While our approach requires some additional manual annotation of the demonstrations, ultimately the approach allows for robust operation on both hierarchical levels: We utilize an optical flow-based method for both computing the similarity between live frame and partial demonstration as well as for estimating the relative 3D transformation of the end-effector. This renders the approach insensitive to view or illumination changes.




With our work, we make the following main contributions: 
\begin{enumerate}
    \item We present a novel framework for sample-efficient recombination of demonstrations on \textit{demonstration graphs}.
    \item We demonstrate the applicability and effectiveness of our method in a number of simulation and real-world experiments, outperforming standard visual servoing methods by a significant margin.
    \item We make our code publicly available at \mbox{\url{http://compservo.cs.uni-freiburg.de}}
\end{enumerate}
 
\section{Related Work}
\label{sec:related_work}

This work aims to realize few-shot imitation in combination with long-horizon planning. Few-shot imitation has been an active area of research, as it greatly increases the practical usability of a robot system. This problem is addressed in several recent works on few-shot imitation, which include DOME \cite{Valassakis2022} and Watch-and-Match \cite{Haldar2022}. Imitation with a focus on category-level imitation has been proposed \cite{Wen2022} and the effectiveness of pre-trained representations has been investigated \cite{Pari2021, nair2022r3m}. Other works employ a servoing approach to imitation \cite{Argus2020, Izquierdo2022}. Few-shot imitation is often realized via goal state conditioning \cite{Pathak2018}. However, these approaches are often designed to only solve single tasks.

Reinforcement learning (RL) has been a popular approach for learning manipulation policies and is often combined with demonstrations to increase learning-wise sample efficiency~\cite{Andrychowicz2017HindsightER, Rajeswaran2017LearningCD}.
As monolithic policies can be unwieldy, there is a line of work into compositionality of RL policies for imitation applied to robot manipulation~\cite{Mandlekar2020, Xu2017, saycan2022}, including ones that focus on representation learning~\cite{Wang2022}.
Furthermore, CompILE~\cite{Kipf2018} analyses compositionality in a more abstract manner.
A multitask benchmark that specifically enables experimenting with transfer between tasks is presented in Meta-world~\cite{Yu2019}. A number of works address the problem of learning policies based on rather unstructured play data \cite{Hangl2020, rosete2022}. Policies can also be improved with prior experience in the form of skill-priors~\cite{Pertsch2020}. An approach augmenting a learned policy with memory from within one episode has been applied to the problem of navigation~\cite{Mezghani2021}.
SayCan~\cite{saycan2022} presents a hierarchical planning algorithm that relies on large-scale language models for high-level planning. Similar to our approach, policies are chosen based on a value function that matches current states to policy options, these are called policy affordances.

Graph-based representations are a well-established approach to solving planning problems \cite{LaValle2006}. For manipulation, this problem can be addressed from a symbolic perspective using task graphs \cite{Huang2018} or programs acted on by graph neural networks \cite{Haro2022}. Planning based on representations obtained by self-supervised learning is also possible as shown in ELAST \cite{gieselmann2022} which applies this to manipulation. Graph-based approaches are very common in the field of robot navigation \cite{Faverjon1996, LaValle1998}. The demonstration graph used by our method is very similar to maps used for navigation, finding similar demonstrations mirrors the problem of localization on those maps \cite{Lim12, Williams11}, which is closely related to the problem of loop closure.

Similar to our work, PALMER \cite{Beker2022} builds a planning graph for the problem of robot navigation. They make use of a hierarchical planning algorithm that is based on a similar strategy of directly using demonstration fragments during inference and relying on an embedding function to identify similar states during planning. Unlike our work, they learn a state embedding function via offline RL. FLAP \cite{Fang2022GeneralizationWL} is also similar and utilizes the graph structure of manipulation problems, but does so by generating embeddings for intermediate states.

Retrieval-based methods are used in a wide variety of settings where data efficiency is important These include few-shot classification~\cite{Parisot2021LongtailRV}, out-of-distribution detection~\cite{sun2022knn}, 3D reconstruction~\cite{what3d}, and image synthesis~\cite{Qi2018SemiParametricIS, Ashual2022KNNDiffusionIG}. Retrieval-based approaches allow direct access to training samples during inference. This essentially constitutes a relaxation of the notion that a policy must be encoded in terms of parameters, or more abstractly as providing a short-cut for information to flow from the training set to inference. This is in line with a number of recent advances in deep learning, such as the addition of skip connections in U-Net \cite{Ronneberger2015UNetCN} and ResNets \cite{He2016IdentityMI} or attention in transformers reducing path-lengths for gradients in sequential computations \cite{Hochreiter2001GradientFI, Vaswani2017AttentionIA}.

\section{Technical Approach}
\label{sec:method}

\begin{figure*}
\centering
 \centering
 \includegraphics[width=.98\linewidth]{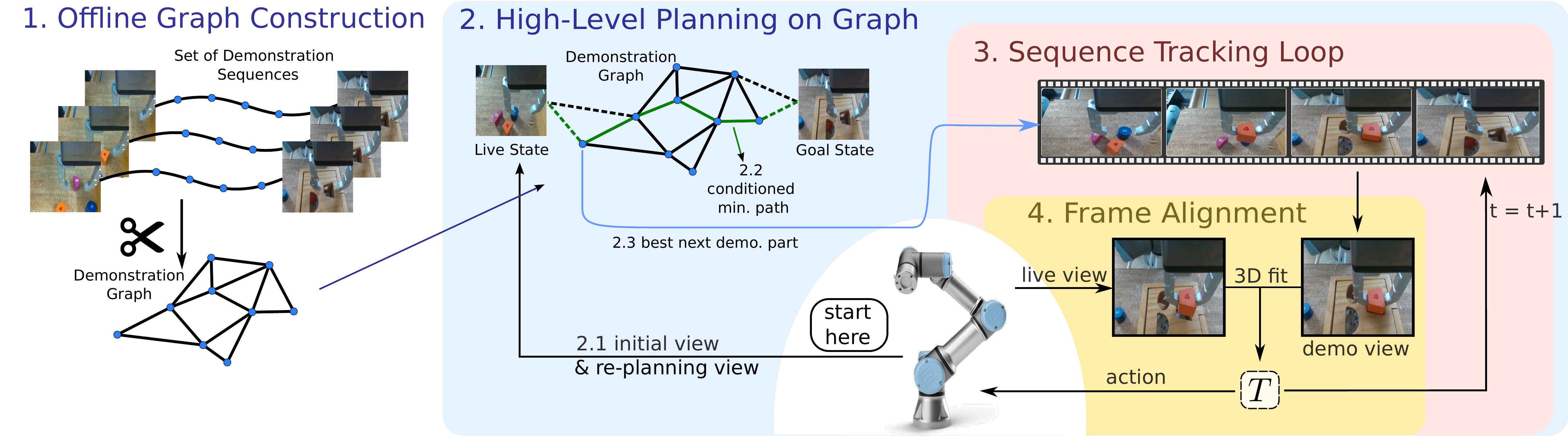}
 \caption{An initial set of demonstrations is segmented into parts and connected on a demonstration graph. The graph edges are weighted based on similarity of the respective start and end frames of the involved trajectories. During inference, we find the minimum-cost path from the current live observation to a given goal observation. Based on the selected demonstration, pixel-wise correspondences between the demo view and the live view are computed via masked optical flow including depth information. The relative 3D transformation of the end-effector is found via least-squares estimation. Sequence tracking queries actions until the magnitude of the relative transform is below a threshold.
 }
 \label{fig:method}
 \vspace{-1em}
\end{figure*}

An environment consists of sets of states $\mathcal{S}$, actions $\mathcal{A}$, and a transition function $p(s', r|s, a)$. Interacting with the environment will result in tuples of $(s, a, r, s')$. We consider successful task execution, which is indicated by a given goal state $s_g$. In this setting, we are given a set of demonstrations $\mathcal{D}$. These are first manually segmented into parts
$M=(s_0,a_0),...,(s_{f},a_{f})$ and stored in a memory bank $\mathcal{M}$. Parts can cover various stages of the task, e.g. localizing an object, grasping an object, or placing it according to a specific task and are then annotated with the pixel segmentation of the foreground object\footnote{Foreground objects are those relative to which the robot is moving. Annotations are also referred to as conditioning $\mathcal{C}$, as it influences servoing.}.

Assuming imitation of parts is possible (see section \cref{ssec:flow_control}), we need to pick parts according to a policy $\pi_\mathcal{M}$, where imitation of picked parts is likely to be successful and which will lead us to the given goal state $s_g$.
Given this setting, we argue that imitation success can be approximately predicted by comparing appearance and position differences of foreground objects. 



\subsection{Graph Based Planning}
\label{ssec:graph_based_planning}
In the simple case, where just one demonstration part is required, the policy $\pi_\mathcal{M}$ should select a part with an initial part state $s_0$ similar to the current state $s_t$ and a final part state $s_f$ similar to the desired goal state $s_g$. In the more complex case, where the concatenation of several demonstration parts is necessary, these can be evaluated by using similarity scores between the last frame of the first part and the first frame of the second part, and so on for more parts.

We quantify the similarity between two states using a function $sim(s^a, s^b) \in \mathbb{R}$. This allows constructing a graph with nodes that are demonstration parts and edge weights given by the similarity function, as indicated in \cref{fig:method}. Adding $s_t$ and $s_g$ to (all nodes of) this graph allows addressing the planning problem by finding the shortest path between these states on the graph.
A proposed path contains an n-tuple of similarity scores $\boldsymbol{s_i}$. In order to obtain a scalar per path, scores can be normalized to the range of $[0,1]$, allowing them to be treated as pseudo-probabilities and combined by multiplication. This is done for all experiments, except for the cross-task experiment  displayed in \cref{fig:multi_task_res}, where treating scores as distances is tested by inverting them before summing.

The planning stage makes the assumption that servoing with respect to a chosen path will lead us to the final state of that demonstration $s_f$, or a very similar state. In practice, this assumption does not always hold. In order to give the algorithm a chance to recover from accumulated errors, a new planning step can be instantiated from the latest observation $s_t$ that was reached via the last servoing step.

In combination, this results in a policy conditioned on a goal state $s_g$, a similarity function $sim(\cdot,\cdot)$, and a set of demonstration parts: $\pi_\mathcal{M}(s_t|s_g, sim, \mathcal{M})$. 
While this may initially sound like a complication of the problem, it comes with a number of advantages:
a) the policy is evaluated in an online fashion using demonstrations; no learning is needed to distill knowledge into a policy, 
b) optimal policies can be pursued based on similar demonstrations, and
c) long-range planning is possible using established path finding algorithms.

\subsection{Similarity Scores}
\label{ssec:sim_scores_method}

To date, there is no evident way to define the similarity function $sim(\cdot,\cdot)$ used for picking suitable demonstrations given $s_t$ and $s_g$. We evaluate several options based on optical flow-based re-projection \cite{Argus2020}, self-supervised feature learning \cite{Pari2021}, contrastive pre-training\cite{nair2022r3m}, and localization \cite{sarlin2019coarse}. These are further described in \cref{ssec:similarity_scores_exp}.

For the scope of this work, we make the assumption that similar appearances and positions of objects correlate with suitable choices.
This may not always hold, in which case learning a similarity function based on task successes may become necessary.
These are not investigated in this work, as we focus on few-shot imitation, in addition to this not relying on training data also reduces the risk of over-fitting on a specific data distribution. 


\subsection{Servoing: Frame Alignment and Sequence Tracking}
\label{ssec:flow_control} 
Once a good demonstration part has been selected by the planning algorithm, the next step is to imitate this part. For this purpose we make use of the FlowControl \cite{Argus2020} algorithm, which uses an optical flow algorithm to find correspondences between the demonstration frames and the live observations.  The algorithm uses demonstrations that consist of RGB-D observations in combination with a manually specified foreground object segmentation that specifies the shape that we want to move relative to. Servoing itself then consists of two steps: frame alignment and sequence tracking.

\textit{Frame alignment} is the process of estimating (and acting upon) the relative transformation between the current scene and the demonstration frame. For this, pixel correspondences between the scenes are obtained from the masked optical flow and then converted into 3D point correspondences using depth information. 
The relative transformation is finally estimated via least-squares and converted into controller instructions.
\textit{Sequence tracking} successively queries actions from the frame alignment routine and applies them until the magnitude of the relative transformation w.r.t.\ the target image is smaller than a certain threshold. Once the tracking process has converged, the next target state is selected.




\section{Experimental Evaluation}
\label{sec:experimental_eval}

\begin{figure}[tb]
    \centering
    \includegraphics[width=.8\linewidth]{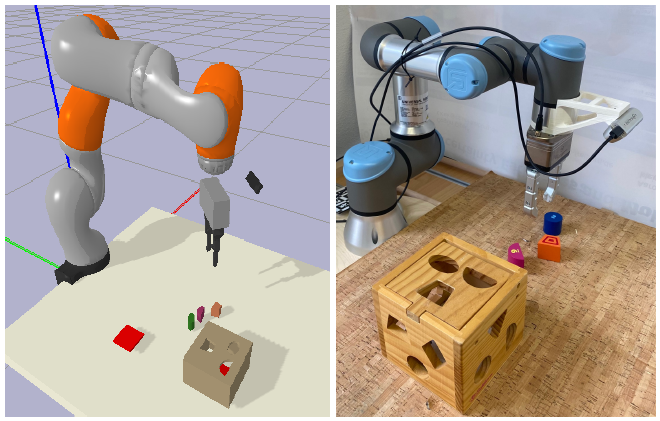}
    \caption{Experimental Setup: In simulation (left) and with the real robot (right). The robot is equipped with an end-of-arm camera that captures images of the scene and is used for recording demonstrations as well as imitating them.}
    \label{fig:setup}
    \vspace{-0.5cm}
\end{figure}

In this section, we evaluate our proposed method on two related manipulation tasks. Similar to previous works \cite{Levine15, Ibarz2021}, we perform shape-sorting, which requires a robot to grasp certain shapes and insert them into a shape-sorting cube. In addition, we evaluate on the easier pick-and-place task, in which the same shapes need to be grasped and then put to a dedicated location without requiring an exact final pose. Hence, the two tasks share a number of intermediate steps, which allows sharing the demonstrations of both tasks as well. In order to demonstrate the efficacy of our approach, the tasks are evaluated both in simulation and on a real robot. The baselines we compare against are FlowControl~\cite{Argus2020}, a visual servoing approach using a single demonstration, and Izquierdo \textit{et al.}~\cite{Izquierdo2022}, which selects an entire demonstration from a set of multiples, without relying on segmented demonstrations.

\subsection{Experimental Setup}
\label{ssec:exp_setup}
As shown in \cref{fig:setup}, the simulation and real-world environments both feature robots in a tabletop scenario with end-of-arm cameras to be used for servoing. Both tasks are simulated in a custom environment using the PyBullet physics simulator \cite{Coumans2021}. In simulation, we use a KUKA iiwa robot, whereas a Universal Robot UR-3 with an end-effector mounted Intel RealSense D435 RGB-D is used in our real-world experiments. For demonstration collection, we make use of a scripted policy in simulation and a 3D Space Mouse for the real-world experiments. 

\subsection{Similarity Scores}
\label{ssec:similarity_scores_exp}
In the following, we list the different similarity scores evaluated in this work, and the methods they are based on.

\noindent\textit{FS: Flow-based similarity:}
We estimate both appearance and position similarity using the optical flow between RGB images. The correctness of the flow is evaluated using the photometric reconstruction as the evaluation function, as done in self-supervised flow estimation methods~\cite{jason2016back, ren2017unsupervised}.
The L2 reprojection distance $d_{RP}$ can be obtained by warping a live view $I^t$ into the demo view $I^d$ using the optical flow, $(u, v)$, between the demo view and the live view. In our case, the difference is evaluated in the region of the demonstration's foreground mask, $\Omega_M$, as the remaining pixels are not relevant for servoing.
\begin{equation}
    d_\text{RP} = \frac{\sum_{x,\, y\,\in\,\Omega_M}{\parallel I^t(x+u,\, y+v) - I ^d(x,\, y)\parallel }}{|~\Omega_M|}
\end{equation}
\begin{equation}
    d_\text{MF} = \frac{\sum_{x,\, y\,\in\,\Omega_M}{\parallel (u, v)\parallel }}{|~\Omega_M|}
\end{equation}
\begin{equation}
\label{eq:sim_fs}
sim_{FS} = d_{RP} + k \cdot d_{MF} 
\end{equation}


In addition, we compute the mean flow $d_{MF}$ by averaging over the flow vectors in the masked part of the image. Finally, we obtain flow-based similarity scores $sim_{FS}$ based on the weighted sum given in \cref{eq:sim_fs} with $k=0.5$.




\noindent\textit{VINN: Visual Imitation through Nearest Neighbors ~\cite{Pari2021}:}
Pari \textit{et al.} define a self-supervised approach to learn the similarity between images. The network is trained by learning feature representations that are invariant to augmentations.
As this method encodes whole images and can be susceptible to distractors, we designed a variant named VINN-Masked, which only uses features from the masked area of the demonstration image to emphasize the focus on the foreground object.

\noindent\textit{R3M: Re-usable Representations for Robot Manipulation~\cite{nair2022r3m}:} 
Nair \textit{et al.} define a learning-based method to generate feature representations that can be used for downstream tasks as a frozen block in the network. In our work, we generate these feature representations from demonstration images. These features are compared to define the most suitable choice of demonstrations for a live episode. Similar to VINN, we used a masked representation here as well, which we define as R3M-Masked.

\noindent\textit{HLOC: Robust hierarchical localization at large scale ~\cite{sarlin2019coarse}: } Sarlin \textit{et al.} first retrieve the best candidates by using global matching descriptors, and then estimate pixel correspondences by using local features.
In our work, we use Superpoint~\cite{SuperPoint} to find 2D pixel correspondences from  masked regions of the demonstration image and the full live image.
Finally, depth values are used to estimate the best 3D fit with RANSAC. The most suitable choice of demonstration is decided by selecting the one with the highest number of inlier points.



\subsection{Simulation Experiments}
\label{sec:exp_sim}

We conduct experiments in simulation to investigate the partitioning of demonstrations, the goal conditioning, and the ability of our approach to transfer demonstrations among different tasks. For simplicity, all simulation experiments use our optical flow-based FS function due to its simplicity and robustness. Trained also on simulated data, the proposed FS function is especially robust to low-texture settings in simulation.



\begin{figure}[tb]
    \centering
    \includegraphics[width=0.75\linewidth,trim={0.0cm 0 0cm 0},clip]{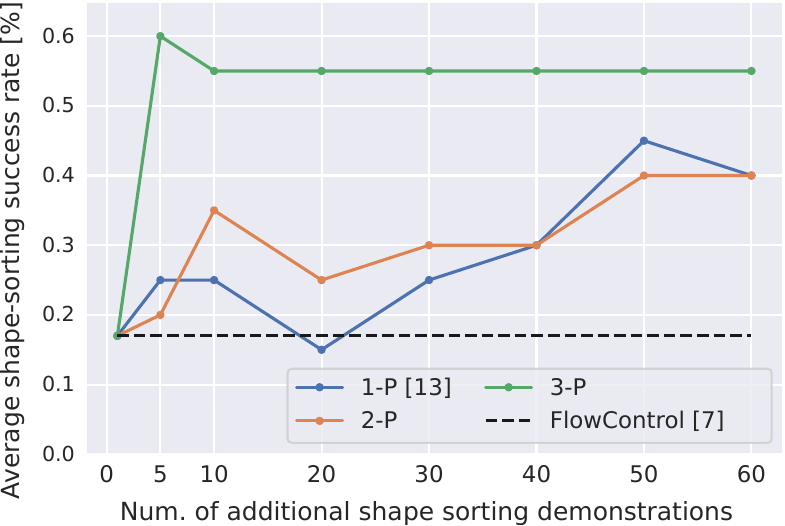}  
    \vspace{-0.2cm}
    \caption{Multi-Part Demonstrations: The shape-sorting performance increases both with more demonstration parts (2-P, 3-P) and an increasing number of demonstrations. The dashed line shows average performance of FlowControl~\cite{Argus2020} with a single randomly picked demonstration.}
    \label{fig:single_task_res}
    \vspace{-1em}
\end{figure}

\textit{Multi-Part Demonstration Graph Search}:
In general, we postulate that splitting demonstrations into multiple parts helps in increasing the servoing success. Thus, in between parts, re-evaluation takes place and the optimal sub-policy given the goal specification is retrieved. We utilize online re-planning, as the best demonstration parts are selected based on live observations. In order to evaluate this hypothesis, we apply the proposed servoing framework using sets of differently partitioned demonstrations as described in \cref{sec:method}. The respective sets contain demonstrations that are either split into one part (1-P), two parts (2-P), or three parts (3-P). The task considered for these experiments is shape-sorting. 
In the 2-P case, the first part deals with locating and correctly orienting the gripper for an optimal grasp, whereas the second part demonstrates the insertion of the shape into the cube. In the 3-P case, the parts guide shape location, gripper re-orientation, and shape insertion, respectively.
As shown in  \cref{fig:single_task_res}, we compare the mean shape-sorting success rates given a certain number of demonstrations across the different demonstration sets (1-P, 2-P, 3-P). The 1-P case corresponds to \cite{Izquierdo2022}, whereas the other baseline using only a single randomly picked demonstration is the FlowControl \cite{Argus2020} method. Overall, we observe that a greater number of demonstrations in all types of demonstration sets on average leads to greater mean shape-sorting success rates. Except for the 1-P case and 20 demonstrations, a higher number of demonstrations always outperforms the single demonstration baseline~\cite{Argus2020}. Thus, using multiple demonstrations allows exploiting a larger pool of available target states. Furthermore, the 3-P case consistently shows higher success rates compared to the 2-P and 1-P experiments.
Especially in the shape-sorting task, it is favorable to reorient the gripper to grasp the object under the correct orientation.
The 3-P experiment includes an additional step that allows selection of the correct demonstration for gripper re-orientation, thereby decoupling the planning of the gripper re-orientation. This increases the probability of overall success even with a reduced pool of demonstrations with sizes as small as 5 demonstrations.


\textit{Goal-Conditioned Demonstration Graph Search}: In the following, we evaluate to what extent goal-conditioning helps in selecting the optimal demonstration to follow. Thus, the initial live frame $s_t$ is compared to potential first demonstration frames $s_0$, while the goal state $s_g$ is compared to a set of terminal demonstration frames $s_f$. We choose a pick-and-place task with three different objects and record demonstrations for two different shapes: trapezes and ovals. In each episode, the robot gripper shall grasp the correct shape and place it on the red surface in order to receive a positive reward, as shown in \cref{fig:setup}.

\begin{table}[t]
    \scriptsize
    \centering
    \caption{Goal-Conditioned Demonstration Graph Search}
    \label{tab:multi_shape}
    \setlength\tabcolsep{5.0pt}
    \begin{threeparttable}
        \begin{tabular}{lc|ccc}
            \toprule
                \multirow{2}{2.0cm}{Demonstrations provided for} & \multirow{2}{1.5cm}{\centering Goal Specification} & \multicolumn{3}{c}{Success Rate [\%]}\\
                & & Trapeze & Oval & Mean \\
                \midrule
                Trapeze or Oval & - & 60 & 80 & 70 \\
                Trapeze \& Oval & - & 45 & 35 & 40\\
                Trapeze \& Oval & \checkmark & \textbf{70} & \textbf{85} & \textbf{78} \\
                \bottomrule
        \end{tabular}
    \footnotesize
    \vspace{.3em}
    \hspace{-.11\linewidth}
    \begin{minipage}{1.21\linewidth}
    To test goal conditioning we use a pick-and-place experiment focusing on two shapes: trapeze and oval, where, in order to obtain success rates, the correct shape must be placed. The demonstration set contains 20 demonstrations for each shape, success rates are computed over 20 trials.
    Using a goal image results in increased performance on the task.
    \end{minipage}
    \end{threeparttable}
    \vspace{-0.3cm}
\end{table}

As given in \cref{tab:multi_shape}, we perform three distinct experiments. In the first experiment, either only demonstrations for the trapeze or the oval shape are provided, while no goal is specified. The second experiment provides demonstrations for both shapes while the goal is still undefined. Different to the second experiment, a goal frame is given for comparison in the third experiment. We observe that a specified goal is crucial to demonstration selection in scenarios with multiple shapes.
This indicates the potential of sharing demonstration parts among a diverse set of tasks.


\begin{figure}[h!]
    \centering
    \includegraphics[width=0.75\linewidth,trim={0cm 0 0cm 0},clip]{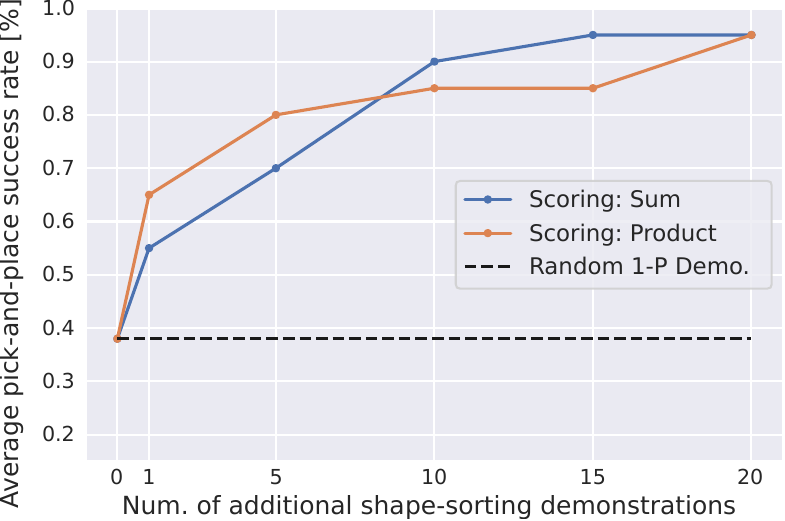}
    \vspace{-0.2cm}
    \caption{Cross-Task Generalization: A higher number of 3-P shape-sorting demonstrations helps in increasing the success on the pick-and-place task, which constitutes the successful transfer of skills between two tasks. The black dashed line shows the averaged performance of FlowControl \cite{Argus2020} that usesa asingle randomly picked 1-P demonstration.}
    \label{fig:multi_task_res}
\end{figure}

\textit{Cross-Task Demonstration Graphs:} 
In this section, we explore the efficacy of combining sets of demonstrations of different tasks, studying the generalization  effects of having a database of demonstrations from a larger domain.
We investigate whether it is possible to perform the pick-and-place task under higher success rates by reusing demonstrations of another task. We collect a varying number of demonstrations of the shape-sorting task and a single demonstration of the pick-and-place task. In addition, a goal image for the pick-and-place task is provided.


Similar to the multi-part scenario in \cref{fig:single_task_res}, we progressively increase the number of shape-sorting demonstrations in the demonstration pool. We provide the experimental results in \cref{fig:multi_task_res}. The average performance of a single randomly sampled \textit{demonstration path} is given as a baseline, with a mean reward of 0.37. While the goal image $s_g$ helps in selecting demonstration parts that place the shape on the corresponding surface, we are able to show that the algorithm is able to use parts of the shape-sorting demonstrations in order to improve the overall pick-and-place task success rate. In general, we observe greater success rates for higher numbers of shape-sorting demonstrations available. Moreover, we notice that the success rates obtained are higher than the mean reward obtained while using a single demonstration of the pick-n-place task.
Although only a limited number of task-respective pick-and-place demonstrations are provided, cross-task demonstrations can be leveraged to increase task success. This effectively demonstrates that skill transfer for visual servoing is feasible and further reiterates on the effectiveness of splitting demonstrations and recombining them optimally.



\begin{figure}[!htb]
     \vspace{-0.2cm}
     \centering
     \includegraphics[width=0.66\linewidth,trim={0 0 0 0},clip]{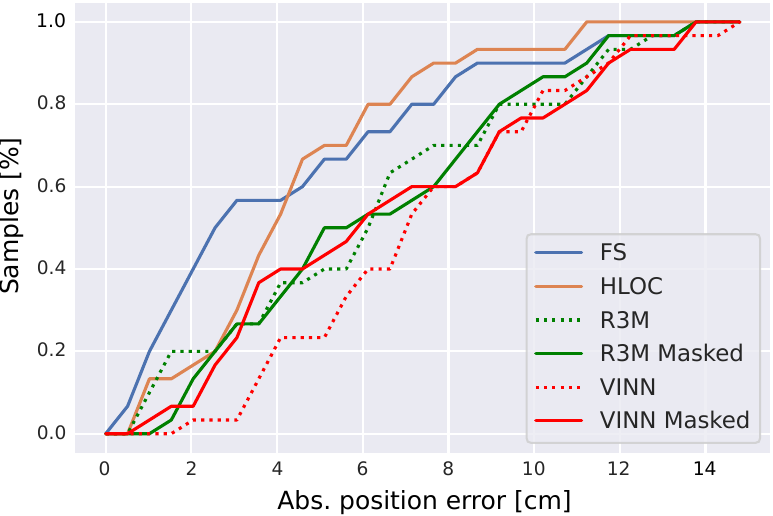}
     \vspace{-0.3cm}
     \caption{Position errors between the most suitable demonstration and a query, computed from initial frames. We compare different methods of selecting the best demonstration and observe that the FS function shows the smallest error.}\label{fig:pos_errors}
\end{figure}

\begin{figure}[!htb]
    \vspace{-0.5cm}
     \centering
     \includegraphics[width=0.66\linewidth,trim={0.0cm 0.0cm 0.0cm 0.0cm},clip]{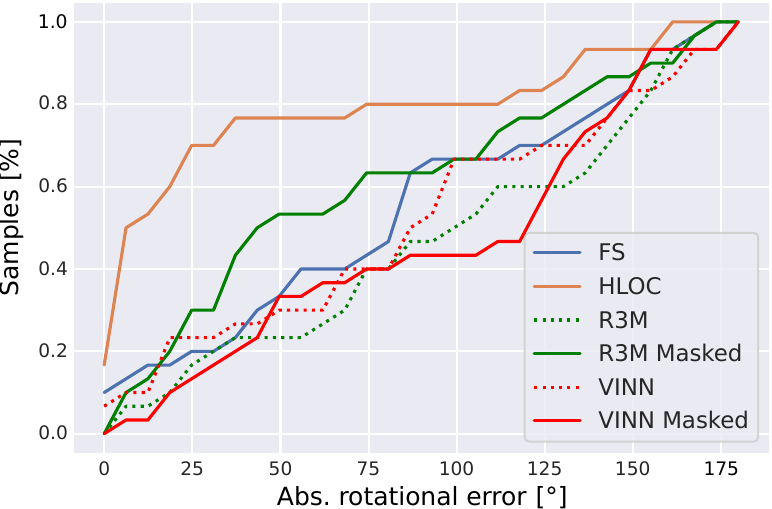}
    \vspace{-0.3cm} 
    \caption{Orientation errors between the most suitable demonstration and a query, computed from initial frames. We compare different methods of selecting the best demonstration and observe that HLOC minimizes the error. }\label{fig:orn_errors}
    \vspace{-.3cm}
\end{figure}

\subsection{Real Robot Evaluation of Similarity Functions}
Using the real-world experimental setup shown in \cref{fig:setup}, we first evaluate the performance of similarity functions based on: FS, HLOC, VINN, and R3M, as described  in \cref{ssec:similarity_scores_exp}.

To obtain observations with randomized object positions, but known object pose differences, we record 30 demonstrations in which objects are grasped with the same relative gripper-object pose\footnote{This is approximate, as done manually, but works because gripper-object pose variance is much smaller than object pose variance between demons.}. The proxy-ground truth pose differences for the initial frame can then be computed by comparing the distances between the robot trajectories to the same (relative) grasping pose. 
Next, we make use of the similarity function to pick frames with estimated \textit{closest} object positions. These values are then compared to the computed distances to evaluate the similarity function.


We report the errors in positions and orientations of the object in the first frame of the query and the best chosen demonstration. These errors are a proxy for how well servoing will work.
The errors regarding position and orientation are displayed in \cref{fig:pos_errors} and \cref{fig:orn_errors}, respectively. We observe that the FS function shows the least position error for small displacements, with HLOC being a close second, while better at larger displacements. In terms of orientation, HLOC shows the least orientation error. This is in line with our expectations, as optical flow is trained with relatively small displacements and rotations compared to localization methods. Considering these results, we conduct additional real robot experiments and report scores using the two strongest similarity functions, i.e., FS and HLOC.




\subsection{Real Robot Experiments} 

In order to verify real-world applicability of the proposed demonstration graphs, we run the multi-part demonstration experiment on the shape-sorting task as described in \cref{sec:exp_sim} on a real-world setup (\cref{fig:setup}). 
Here, we test the 1-P and the 3-P configurations while using the FS and HLOC similarity functions, to show that these can be replaced in a modular manner. The results of this experiment are shown in \cref{tab:sr_real}.
While using the FS function, the 1-P setting is successful in 2/20 cases (10\%). The 3-P graphs show 10/20 (50\%) successful runs. These results are similar to those obtained in the simulation shown in \cref{fig:single_task_res}.
In comparison, HLOC is successful in 11/20 (55\%) and 17/20 (85\%) cases in the 1-P and 3-P graph settings, respectively.

During evaluation, we report intermediate success rates for correct localization, grasping, alignment, and insertion so that the shape-sorting performance can be evaluated along the different task stages. We observe that in all cases, the highest performance drop occurs during grasping. Nonetheless, the 3-P HLOC-based method is capable of executing 90\% of all grasps. Above all, this shows that similarity functions based on optical flow and localization networks allow zero-shot application in real-world settings. 


    
\begin{table}[tb]
    \scriptsize
    \centering
    \caption{Real-Robot Shape-Sorting}
    \label{tab:sr_real}
    \setlength\tabcolsep{3.0pt}
    \begin{threeparttable}
        \begin{tabular}{ll|cccc|c}
        \toprule
            \multirow{2}{1.4cm}{Demonstration Partitioning} & \multirow{2}{1.0cm}{Similarity Function} &  \multirow{2}{0.7cm}{\centering Initial Runs } & \multirow{2}{0.8cm}{\centering Correct Position} & \multirow{2}{0.8cm}{\centering Correct Grasp} & \multirow{2}{1.1cm}{\centering Correct Orientation} & \multirow{2}{0.8cm}{\centering Success} \\
             &  &  &  &  & & \\
            \midrule
            1-P \cite{Izquierdo2022}       & FS  &  20 & 17 & 7 & 5 & 2 \\
            3-P (ours) & FS  &  20 & 18 & 16 & 13 & 10 \\
            \midrule
            1-P \cite{Izquierdo2022} & HLOC \cite{sarlin2019coarse} & 20 & 20 & 13 & 13 & 11 \\
            3-P (ours) & HLOC \cite{sarlin2019coarse} &  20 & \textbf{20} & \textbf{18} & \textbf{18}& \textbf{17} \\
            \bottomrule
        \end{tabular}
        \footnotesize
        Real-robot experiment evaluated on the shape-sorting task with 20 demonstrations.
        We evaluate the 1-P and the 3-P case.
        Additionally, we evaluate the success rate at each stage in the algorithm.
        The highest drop in the success rate occurs at the grasping stage because of the inability of the gripper to re-orient itself correctly.
    \end{threeparttable}
    \vspace{-0.4cm}
\end{table}

\section{Conclusion}

This work presents a framework for few-shot imitation that involves segmenting and recombining partial demonstrations in a combinatorial manner. This allows our approach to show improved performance on numerous tasks under varying environment conditions compared to standard visual servoing baselines. Ultimately, hierarchical planning on the introduced demonstration graphs enables multi-modal long-horizon manipulation. Planning is based on similarity functions which can be used in a modular way.
In addition, the use of optical flow-based correspondence estimation guarantees robustness to visual and scene-wise variations. The proposed method was evaluated extensively in both simulation and real-world settings, demonstrating the practical applicability.



\footnotesize
\bibliographystyle{IEEEtran}
\bibliography{references.bib}

\begin{thebibliography}{10}
\providecommand{\url}[1]{#1}
\csname url@rmstyle\endcsname
\providecommand{\newblock}{\relax}
\providecommand{\bibinfo}[2]{#2}
\providecommand\BIBentrySTDinterwordspacing{\spaceskip=0pt\relax}
\providecommand\BIBentryALTinterwordstretchfactor{4}
\providecommand\BIBentryALTinterwordspacing{\spaceskip=\fontdimen2\font plus
\BIBentryALTinterwordstretchfactor\fontdimen3\font minus \fontdimen4\font\relax}
\providecommand\BIBforeignlanguage[2]{{%
\expandafter\ifx\csname l@#1\endcsname\relax
\typeout{** WARNING: IEEEtran.bst: No hyphenation pattern has been}%
\typeout{** loaded for the language `#1'. Using the pattern for}%
\typeout{** the default language instead.}%
\else
\language=\csname l@#1\endcsname
\fi
#2}}

\bibitem{breazeal2002robots}
C.~Breazeal and B.~Scassellati, ``Robots that imitate humans,'' \emph{Trends in cognitive sciences}, vol.~6, no.~11, pp. 481--487, 2002.

\bibitem{celemin2022interactive}
C.~Celemin, R.~P{\'e}rez-Dattari, E.~Chisari, G.~Franzese, L.~de~Souza~Rosa, R.~Prakash, Z.~Ajanovi{\'c}, M.~Ferraz, A.~Valada, J.~Kober, \emph{et~al.}, ``Interactive imitation learning in robotics: A survey,'' \emph{Foundations and Trends{\textregistered} in Robotics}, vol.~10, no. 1-2, pp. 1--197, 2022.

\bibitem{Valassakis2022}
E.~Valassakis, G.~Papagiannis, N.~D. Palo, and E.~Johns, ``Demonstrate once, imitate immediately (dome): Learning visual servoing for one-shot imitation learning,'' \emph{Int.~Conf.~on Intelligent Robots and Systems}, pp. 8614--8621, 2022.

\bibitem{zhang2018deep}
T.~Zhang, Z.~McCarthy, O.~Jow, D.~Lee, X.~Chen, K.~Goldberg, and P.~Abbeel, ``Deep imitation learning for complex manipulation tasks from virtual reality teleoperation,'' in \emph{Int.~Conf.~on Robotics and Automation}, 2018, pp. 5628--5635.

\bibitem{finn2017one}
C.~Finn, T.~Yu, T.~Zhang, P.~Abbeel, and S.~Levine, ``One-shot visual imitation learning via meta-learning,'' in \emph{Conference on robot learning}, 2017, pp. 357--368.

\bibitem{Mitsuda99}
T.~Mitsuda, Y.~Miyazaki, N.~Maru, K.~F. MacDorman, A.~Nishikawa, and F.~Miyazaki, ``Visual servoing based on coarse optical flow,'' \emph{IFAC Proceedings Volumes}, vol.~32, no.~2, pp. 539 -- 544, 1999.

\bibitem{Argus2020}
M.~Argus, L.~Hermann, J.~Long, and T.~Brox, ``Flowcontrol: Optical flow based visual servoing,'' \emph{Int.~Conf.~on Intelligent Robots and Systems}, pp. 7534--7541, 2020.

\bibitem{Parisot2021LongtailRV}
S.~Parisot, P.~M. Esperança, S.~G. McDonagh, T.~Madar{\'a}sz, Y.~Yang, and Z.~Li, ``Long-tail recognition via compositional knowledge transfer,'' \emph{Proc.~of the IEEE Conf.~on Computer Vision and Pattern Recognition}, pp. 6929--6938, 2021.

\bibitem{sun2022knn}
Y.~Sun, Y.~Ming, X.~Zhu, and Y.~Li, ``Out-of-distribution detection with deep nearest neighbors,'' in \emph{Int.~Conf.~on Machine Learning}, 2022, pp. 20\,827--20\,840.

\bibitem{what3d}
M.~Tatarchenko, S.~R. Richter, R.~Ranftl, Z.~Li, V.~Koltun, and T.~Brox, ``What do single-view 3d reconstruction networks learn?'' in \emph{Proc.~of the IEEE Conf.~on Computer Vision and Pattern Recognition}, 2019.

\bibitem{Qi2018SemiParametricIS}
X.~Qi, Q.~Chen, J.~Jia, and V.~Koltun, ``Semi-parametric image synthesis,'' \emph{Proc.~of the IEEE Conf.~on Computer Vision and Pattern Recognition}, pp. 8808--8816, 2018.

\bibitem{Ashual2022KNNDiffusionIG}
O.~Ashual, S.~Sheynin, A.~Polyak, U.~Singer, O.~Gafni, E.~Nachmani, and Y.~Taigman, ``Knn-diffusion: Image generation via large-scale retrieval,'' \emph{ArXiv}, vol. abs/2204.02849, 2022.

\bibitem{Izquierdo2022}
S.~Izquierdo, M.~Argus, and T.~Brox, ``Conditional visual servoing for multi-step tasks,'' \emph{Int.~Conf.~on Intelligent Robots and Systems}, pp. 2190--2196, 2022.

\bibitem{Haldar2022}
S.~Haldar, V.~Mathur, D.~Yarats, and L.~Pinto, ``Watch and match: Supercharging imitation with regularized optimal transport,'' in \emph{Conf.~on Robot Learning}, 2023, pp. 32--43.

\bibitem{Wen2022}
B.~Wen, W.~Lian, K.~E. Bekris, and S.~Schaal, ``You only demonstrate once: Category-level manipulation from single visual demonstration,'' \emph{Robotics: Science and Systems}, 2022.

\bibitem{Pari2021}
J.~Pari, N.~M.~M. Shafiullah, S.~P. Arunachalam, and L.~Pinto, ``The surprising effectiveness of representation learning for visual imitation,'' \emph{Robotics: Science and Systems}, 2022.

\bibitem{nair2022r3m}
S.~Nair, A.~Rajeswaran, V.~Kumar, C.~Finn, and A.~Gupta, ``R3m: A universal visual representation for robot manipulation,'' \emph{Conf.~on Robot Learning}, 2022.

\bibitem{Pathak2018}
D.~Pathak, P.~Mahmoudieh, G.~Luo, P.~Agrawal, D.~Chen, Y.~Shentu, E.~Shelhamer, J.~Malik, A.~A. Efros, and T.~Darrell, ``Zero-shot visual imitation,'' \emph{Proc.~of the IEEE Conf.~on Computer Vision and Pattern Recognition Workshops}, pp. 2131--21\,313, 2018.

\bibitem{Andrychowicz2017HindsightER}
M.~Andrychowicz, F.~Wolski, A.~Ray, J.~Schneider, R.~Fong, P.~Welinder, B.~McGrew, J.~Tobin, O.~Pieter~Abbeel, and W.~Zaremba, ``Hindsight experience replay,'' \emph{Proc.~of the Conf.~on Neural Information Processing Systems (NIPS)}, vol.~30, 2017.

\bibitem{Rajeswaran2017LearningCD}
A.~Rajeswaran, V.~Kumar, A.~Gupta, G.~Vezzani, J.~Schulman, E.~Todorov, and S.~Levine, ``{Learning Complex Dexterous Manipulation with Deep Reinforcement Learning and Demonstrations},'' in \emph{Robotics: Science and Systems}, 2018.

\bibitem{Mandlekar2020}
A.~Mandlekar, D.~Xu, R.~Mart{\'i}n-Mart{\'i}n, S.~Savarese, and L.~Fei-Fei, ``Learning to generalize across long-horizon tasks from human demonstrations,'' \emph{Robotics: Science and Systems}, 2020.

\bibitem{Xu2017}
D.~Xu, S.~Nair, Y.~Zhu, J.~Gao, A.~Garg, L.~Fei-Fei, and S.~Savarese, ``Neural task programming: Learning to generalize across hierarchical tasks,'' \emph{Int.~Conf.~on Robotics and Automation}, pp. 1--8, 2017.

\bibitem{saycan2022}
M.~Ahn, A.~Brohan, N.~Brown, Y.~Chebotar, O.~Cortes, B.~David, C.~Finn, C.~Fu, K.~Gopalakrishnan, K.~Hausman, A.~Herzog, D.~Ho, J.~Hsu, J.~Ibarz, B.~Ichter, A.~Irpan, E.~Jang, R.~J. Ruano, K.~Jeffrey, S.~Jesmonth, N.~Joshi, R.~Julian, D.~Kalashnikov, Y.~Kuang, K.-H. Lee, S.~Levine, Y.~Lu, L.~Luu, C.~Parada, P.~Pastor, J.~Quiambao, K.~Rao, J.~Rettinghouse, D.~Reyes, P.~Sermanet, N.~Sievers, C.~Tan, A.~Toshev, V.~Vanhoucke, F.~Xia, T.~Xiao, P.~Xu, S.~Xu, M.~Yan, and A.~Zeng, ``Do as i can and not as i say: Grounding language in robotic affordances,'' in \emph{arXiv preprint arXiv:2204.01691}, 2022.

\bibitem{Wang2022}
C.~Wang, D.~Xu, and L.~Fei-Fei, ``Generalizable task planning through representation pretraining,'' \emph{IEEE Robotics and Automation Letters}, vol.~7, pp. 8299--8306, 2022.

\bibitem{Kipf2018}
T.~Kipf, Y.~Li, H.~Dai, V.~F. Zambaldi, A.~Sanchez-Gonzalez, E.~Grefenstette, P.~Kohli, and P.~W. Battaglia, ``Compile: Compositional imitation learning and execution,'' in \emph{Int.~Conf.~on Machine Learning}, 2018.

\bibitem{Yu2019}
T.~Yu, D.~Quillen, Z.~He, R.~Julian, K.~Hausman, C.~Finn, and S.~Levine, ``Meta-world: A benchmark and evaluation for multi-task and meta reinforcement learning,'' in \emph{Conf.~on Robot Learning}, 2020, pp. 1094--1100.

\bibitem{Hangl2020}
S.~Hangl, V.~Dunjko, H.~J. Briegel, and J.~H. Piater, ``Skill learning by autonomous robotic playing using active learning and exploratory behavior composition,'' \emph{Frontiers in Robotics and AI}, vol.~7, 2020.

\bibitem{rosete2022}
E.~Rosete-Beas, O.~Mees, G.~Kalweit, J.~Boedecker, and W.~Burgard, ``Latent plans for task agnostic offline reinforcement learning,'' in \emph{Conf.~on Robot Learning}, Auckland, New Zealand, 2022.

\bibitem{Pertsch2020}
K.~Pertsch, Y.~Lee, and J.~J. Lim, ``Accelerating reinforcement learning with learned skill priors,'' in \emph{Conf.~on Robot Learning}, 2020.

\bibitem{Mezghani2021}
L.~Mezghani, S.~Sukhbaatar, T.~Lavril, O.~Maksymets, D.~Batra, P.~Bojanowski, and A.~Karteek, ``Memory-augmented reinforcement learning for image-goal navigation,'' \emph{Int.~Conf.~on Intelligent Robots and Systems}, pp. 3316--3323, 2021.

\bibitem{LaValle2006}
S.~M. LaValle, \emph{Planning Algorithms}.\hskip 1em plus 0.5em minus 0.4em\relax USA: Cambridge University Press, 2006.

\bibitem{Huang2018}
D.-A. Huang, S.~Nair, D.~Xu, Y.~Zhu, A.~Garg, L.~Fei-Fei, S.~Savarese, and J.~C. Niebles, ``Neural task graphs: Generalizing to unseen tasks from a single video demonstration,'' in \emph{Proc.~of the IEEE Conf.~on Computer Vision and Pattern Recognition}, 2019, pp. 8557--8566.

\bibitem{Haro2022}
J.~Ortiz-Haro, J.-S. Ha, D.~Driess, E.~Karpas, and M.~Toussaint, ``Learning feasibility of factored nonlinear programs in robotic manipulation planning,'' in \emph{Int.~Conf.~on Robotics and Automation}, 2023, pp. 3729--3735.

\bibitem{gieselmann2022}
R.~Gieselmann and F.~T. Pokorny, ``Latent planning via expansive tree search,'' in \emph{Proc.~of the Conf.~on Neural Information Processing Systems (NIPS)}, A.~H. Oh, A.~Agarwal, D.~Belgrave, and K.~Cho, Eds., 2022.

\bibitem{Faverjon1996}
L.~Kavraki, P.~Svestka, J.-C. Latombe, and M.~Overmars, ``Probabilistic roadmaps for path planning in high-dimensional configuration spaces,'' \emph{{IEEE} Tran. on Rob. and Aut.}, vol.~12, no.~4, pp. 566--580, 1996.

\bibitem{LaValle1998}
S.~M. LaValle, ``Rapidly-exploring random trees : a new tool for path planning,'' \emph{The annual research report}, 1998.

\bibitem{Lim12}
H.~Lim, S.~N. Sinha, M.~F. Cohen, and M.~Uyttendaele, ``Real-time image-based 6-dof localization in large-scale environments,'' \emph{Proc.~of the IEEE Conf.~on Computer Vision and Pattern Recognition}, pp. 1043--1050, 2012.

\bibitem{Williams11}
B.~Williams, G.~S.~W. Klein, and I.~Reid, ``Automatic relocalization and loop closing for real-time monocular slam,'' \emph{IEEE Transactions on Pattern Analysis and Machine Intelligence}, vol.~33, pp. 1699--1712, 2011.

\bibitem{Beker2022}
O.~Beker, M.~Mohammadi, and A.~Zamir, ``Palmer: Perception-action loop with memory for long-horizon planning,'' \emph{Proc.~of the Conf.~on Neural Information Processing Systems (NIPS)}, vol.~35, pp. 34\,258--34\,271, 2022.

\bibitem{Fang2022GeneralizationWL}
K.~Fang, P.~Yin, A.~Nair, H.~Walke, G.~Yan, and S.~Levine, ``Generalization with lossy affordances: Leveraging broad offline data for learning visuomotor tasks,'' in \emph{Conf.~on Robot Learning}, 2022.

\bibitem{Ronneberger2015UNetCN}
O.~Ronneberger, P.~Fischer, and T.~Brox, ``U-net: Convolutional networks for biomedical image segmentation,'' in \emph{Int.~Conf. on Med. Image Comp. and Comp.-Assisted Interv.}, 2015, pp. 234--241.

\bibitem{He2016IdentityMI}
K.~He, X.~Zhang, S.~Ren, and J.~Sun, ``Identity mappings in deep residual networks,'' in \emph{Europ.~Conf.~on Computer Vision}.\hskip 1em plus 0.5em minus 0.4em\relax Springer, 2016, pp. 630--645.

\bibitem{Hochreiter2001GradientFI}
S.~Hochreiter and Y.~Bengio, ``Gradient flow in recurrent nets: the difficulty of learning long-term dependencies,'' in \emph{A Field Guide to Dynamical Recurrent Networks}, 2001.

\bibitem{Vaswani2017AttentionIA}
A.~Vaswani, N.~Shazeer, N.~Parmar, J.~Uszkoreit, L.~Jones, A.~N. Gomez, {\L}.~Kaiser, and I.~Polosukhin, ``Attention is all you need,'' \emph{Proc.~of the Conf.~on Neural Information Processing Systems (NIPS)}, vol.~30, 2017.

\bibitem{sarlin2019coarse}
P.-E. Sarlin, C.~Cadena, R.~Siegwart, and M.~Dymczyk, ``From coarse to fine: Robust hierarchical localization at large scale,'' in \emph{Proc.~of the IEEE Conf.~on Computer Vision and Pattern Recognition}, 2019, pp. 12\,716--12\,725.

\bibitem{Levine15}
S.~Levine, C.~Finn, T.~Darrell, and P.~Abbeel, ``End-to-end training of deep visuomotor policies,'' \emph{J. Mach. Learn. Res.}, vol.~17, pp. 39:1--39:40, 2015.

\bibitem{Ibarz2021}
J.~Ibarz, J.~Tan, C.~Finn, M.~Kalakrishnan, P.~Pastor, and S.~Levine, ``How to train your robot with deep reinforcement learning: lessons we have learned,'' \emph{Int.~Journal of Robotics Research}, vol.~40, pp. 698 -- 721, 2021.

\bibitem{Coumans2021}
E.~Coumans and Y.~Bai, ``Pybullet, a python module for physics simulation for games, robotics and machine learning,'' \url{http://pybullet.org}, 2016--2021.

\bibitem{jason2016back}
J.~Y. Jason, A.~W. Harley, and K.~G. Derpanis, ``Back to basics: Unsupervised learning of optical flow via brightness constancy and motion smoothness,'' in \emph{Europ.~Conf.~on Computer Vision}.\hskip 1em plus 0.5em minus 0.4em\relax Springer, 2016, pp. 3--10.

\bibitem{ren2017unsupervised}
Z.~Ren, J.~Yan, B.~Ni, B.~Liu, X.~Yang, and H.~Zha, ``Unsupervised deep learning for optical flow estimation,'' in \emph{Proc.~of the AAAI Conference on Artificial Intelligence (AAAI)}, 2017.

\bibitem{SuperPoint}
D.~DeTone, T.~Malisiewicz, and A.~Rabinovich, ``Superpoint: Self-supervised interest point detection and description,'' \emph{Proc.~of the IEEE Conf.~on Computer Vision and Pattern Recognition Workshops}, pp. 337--33\,712, 2018.

\end{thebibliography}


\end{document}